\documentclass[runningheads,a4paper]{llncs}

\usepackage{multirow}
\usepackage{amssymb}
\usepackage{color}
\usepackage{graphicx}

\usepackage{booktabs}
\usepackage{amsmath}
\usepackage{algorithm}
\usepackage[noend]{algpseudocode}
\usepackage{caption}
\usepackage{rotating}
\usepackage{paralist}
\usepackage{setspace}

\usepackage{url}
\urldef{\mailsa}\path|{tfbt-public@google.com}|
\newcommand{\keywords}[1]{\par\addvspace\baselineskip
\noindent\keywordname\enspace\ignorespaces#1}

\newsavebox\CBox

\begin{document}

\mainmatter  

\title{TF Boosted Trees: A scalable TensorFlow based framework for gradient boosting}

\titlerunning{TFBT: gradient boosting in TensorFlow}

%
%
\author{Natalia Ponomareva, 
Soroush Radpour,
Gilbert Hendry, 
Salem Haykal,
Thomas Colthurst, 
Petr Mitrichev,
Alexander Grushetsky}
\authorrunning{TFBT: gradient boosting in TensorFlow}

\institute{Google, Inc.
\\
\url{tfbt-public@google.com}}

%
%

\toctitle{Lecture Notes in Computer Science}
\tocauthor{Authors' Instructions}
\maketitle

\begin{abstract}
TF Boosted Trees (TFBT) is a new open-sourced framework for the distributed training of gradient boosted trees. It is based on TensorFlow, and its distinguishing features include a novel architecture, automatic loss differentiation, layer-by-layer boosting that results in smaller ensembles and faster prediction, principled multi-class handling, and a number of regularization techniques to prevent overfitting.
\end{abstract}

\section{Introduction}
Gradient boosted trees are deservedly popular machine learning models.
Since their introduction in \cite{friedman99} they have gone on to dominate many competitions on real-world data, including Kaggle and KDDCup \cite{xgboost}. In addition to their excellent accuracy, they are also easy to use, as they deal well with unnormalized, collinear, missing, or outlier-infected data. They can also support custom loss functions and are often easier to interpret than neural nets or large linear models. Because of their popularity, there are now many gradient boosted tree implementations, including scikit-learn \cite{scikit-learn}, R gbm \cite{gbm}, Spark MLLib \cite{Meng:2016:MML:2946645.2946679}, LightGBM \cite{LightGBM} and XGBoost \cite{xgboost}.

In this paper, we introduce another optimized and scalable gradient boosted tree library, \textbf{TF Boosted Trees (TFBT)}, which is built on top of the TensorFlow framework \cite{tensorflow}. TFBT incorporates a number of novel algorithmic improvements to the gradient boosting algorithm, including new per-layer boosting procedure which offers improved performance on some problems.
TFBT is open source, and available in the main TensorFlow distribution  under \verb+contrib/boosted_trees+.

\section{TFBT features}
\label{features-label}
In Table \ref{libs_comparison} we provide a brief comparison between TFBT and some existing libraries. 
Additionally, TFBT provides the following.

\textbf{Layer-by-layer boosting.} TFBT supports two modes of tree building: \textit{standard} (building sequence of boosted trees in a stochastic gradient fashion) and novel \textit{Layer-by-Layer} boosting, which allows for stronger trees (leading to faster convergence) and deeper models.
One weakness of tree-based methods is the fact that only the examples falling under a given partition are used to produce the estimator associated with that leaf, so deeper nodes use statistics calculated from fewer examples.
We overcome that limitation by recalculating the gradients and Hessians whenever a new layer is built resulting in stronger trees that better approximate the functional space gradient.
This enables deeper nodes to use higher level splits as priors meaning each new layer will have more information and will be able to better adjust for errors from the previous layers.
Empirically we found that layer-by-layer boosting generally leads to faster convergence and, with proper regularization, to less overfitting for deeper trees.

\textbf{Multiclass support}. TFBT supports one-vs-rest, as well as other variations that reduce the number of required trees by storing per-class scores at each leaf. All other implementations use one-vs-rest as a multiclass handling strategy (MLLib does not support multiclass at all).

\begin{table}[ht!]
\small
\centering
\caption{Comparison of gradient boosted libraries.}
\scriptsize	
\label{libs_comparison}
\renewcommand{\arraystretch}{1}
\setlength{\tabcolsep}{5pt}
\begin{tabular}{p{0.9cm} p{0.15cm} p{4.9cm} p{5.1cm}}
\toprule
\textbf{Lib} & \textbf{D?}  & \centering \textbf{Losses                                                                                                                                                  } & \textbf{Regularization} \\
\midrule
scikit-learn & N & \textit{R}: least squares, least absolute dev, huber and quantile.
\textit{C}:  logistic, Max-Ent and exp & Depth limit, shrinkage, bagging, feature subsampling  \\
\hline
GBM  & N  & \textit{R}: least squares, least absolute dev, t-distribution, quantile, huber.
\textit{C}:  logistic, Max-Ent, exp, poisson \& right censored observations. 
Supports \textit{ranking} & Shrinkage, bagging, depth limit, min \# of examples per node.  \\
\hline
MLLib & Y & \textit{R}: least squared and least absolute dev.
\textit{C}: logistic. 
& Shrinkage, early stopping, depth limit, min \# of examples per node, min gain, bagging.  \\
\hline
Light GBM & Y & \textit{R}: least squares, least absolute dev, huber, fair, poisson.
\textit{C}: logistic, Max-Ent. Supports \textit{ranking}.  
& Dropout, shrinkage, \# leafs limit, feature subsampling, bagging, L1 \& L2 \\
\hline
XGBoost  & Y & \textit{R}: least squares, poisson, gamma, tweedie regression.
\textit{C}: logistic, Max-Ent. Supports \textit{ranking} and \textbf{custom}. 
& L1 \& L2, shrinkage, feature subsampling, dropout, bagging, min child weight and gain, limit on depth and \# of nodes, pruning. \\
\hline
\textbf{TFBT}  & Y & Any twice differentiable loss from tf.contrib.losses and \textbf{custom} losses. 
& L1 \& L2, tree complexity, shrinkage, line search for learning rate, dropout, feature subsampling and bagging, limit on depth and min node weight, pre-\/ post- pruning.\\
\bottomrule 
 \multicolumn{4}{l}{\tiny \textit{D?} is whether a library supports distributed mode. \textit{R} stands for regression, \textit{C} for classification.} 
\end{tabular}
\end{table}

Since TFBT is implemented in TensorFlow, all of \textbf{TensorFlow specific features} are also available 
\begin{compactitem}
\item Ease of writing \textbf{custom loss} functions, as TensorFlow provides automatic differentiation \cite{tensorflow} (other packages like XGBoost require the user to provide the first and second order derivatives). 
\item Ability to seamlessly switch and compare TFBT and other canned TensorFlow models, as well as ease of composing gradient boosted tree models with features produced by other TensorFlow models. 
\item Ease of debugging with TensorBoard.
\item Models can be run on multiple CPUs/GPUs and on multiple platforms, including  mobile, and can be easily deployed via TF serving \cite{tf-serving}
\item Checkpointing for fault tolerance, incremental training \& warm restart.
\end{compactitem}

\section{TFBT system design}
\vspace*{-0.3cm}
\textbf{Finding splits}. One of the most computationally intensive parts in boosting is finding the best splits. Both R and scikit-learn work with an exact greedy algorithm for enumerating all possible splits for all possible features, which does not scale well. The remaining implementations, including XGBoost, work with approximate algorithms to build quantiles of feature value and aggregating gradients and Hessians for each bucket of quantiles. To aggregate the statistics, two approaches can be used \cite{tencentboost}: either each of the workers works on all the features, and then the statistics are aggregated in
Map-Reduce (like in MLLib) or All-Reduce (XGBoost) fashion, or a parameter server (PS) approach (similar to TencentBoost\cite{tencentboost} and PSMART \cite{psmart}) is applied, where each worker and PS aggregates statistics only for a subset of features. The All-Reduce versions do not scale to a high-dimensional data and Map-Reduce
versions are slow to scale. 

\textbf{TFBT Architecture.} Our computation model is based on the following needs:
\begin{compactenum}
\item Ability to train on datasets that don't fit in workers' memory.
\item Ability to train deeper trees with a larger number of features.
\item Support for different modes of building the trees: standard one-tree-per-batch mode, as well as boosting the tree layer-by-layer.
\item Minimizing parallelization costs. Low cost restarts on stateless workers would allow us to use much cheaper preemptible VMs.
\end{compactenum}

\vspace*{-0.32cm}
\begin{figure}[h!] 
  \caption{TFBT architecture.}
  \centering
  \includegraphics[scale=0.33]{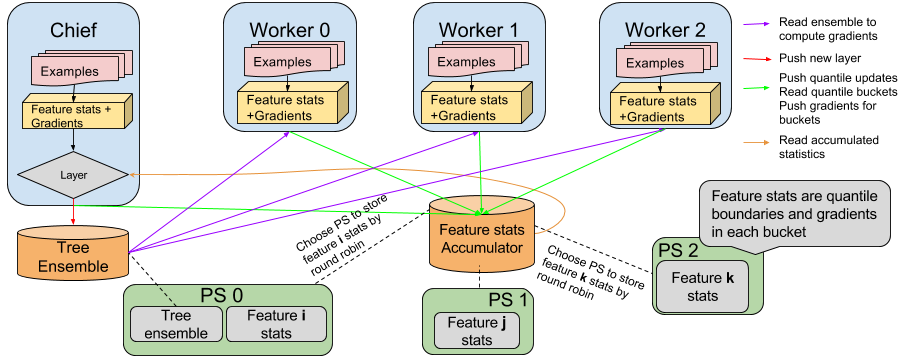}
  \label{figure:architecture}
\end{figure}
 
\makeatletter
\def\BState{\State\hskip-\ALG@thistlm}
\makeatother
\algnewcommand\And{\textbf{and}}
\algrenewcommand\alglinenumber[1]{\scriptsize #1:}
\begin{algorithm}
\scriptsize	
\caption{Chief and Workers' work}
\begin{algorithmic}[1]
\Procedure{CalculateStatistics(ps, model, stamp, batch\_data, loss\_fn)}{}
\State $predictions \gets \mbox{model.predict(BATCH\_DATA)}$
\State $quantile\_stats \gets \mbox{calculate\_quantile\_stats(BATCH\_DATA)}$
\State push\_stats(PS,quantile\_stats, stamp) \Comment{\parbox[t]{.35\linewidth}{PS updates quantiles}} 
\State $current\_boundaries \gets \mbox{fetch\_latest\_boundaries(PS, stamp)} $ 
\State $gradients, hessians \gets \mbox{calculate\_derivatives(predictions,LOSS\_FN)}$
\State $gradients, hessians \gets \mbox{aggregate(current\_boundaries,gradients, hessians)}$
\State push\_stats(PS,gradients, hessians, size(BATCH\_DATA), stamp)
\EndProcedure
\Procedure{DoWork(ps, loss\_fn, is\_chief)}{} \Comment{\parbox[t]{.35\linewidth}{Runs on workers and 1 chief}} 
\While {true}
\State $BATCH\_DATA \gets \mbox{read\_data\_batch()}$
\State $model \gets \mbox{fetch\_latest\_model(PS)}$
\State $stamp \gets \mbox{model.stamp\_token}$
\State CalculateStatistics(PS,model, stamp, BATCH\_DATA,LOSS\_FN)
\If{$is\_chief \And \mbox{get\_num\_examples(PS, stamp)}} \textgreater=N\_PER\_LAYER$
\State $next\_stamp \gets stamp+1$
\State $stats \gets \mbox{flush(PS, stamp, next\_stamp)}$ \Comment{\parbox[t]{.35\linewidth}{Update stamp, returns stats}} 
\State build\_layer(PS, model,  next\_stamp, stats) \Comment{\parbox[t]{.35\linewidth}{PS updates ensemble}} 
\EndIf
\EndWhile
\EndProcedure
\end{algorithmic}
\label{Algo}
\end{algorithm}

Our design is similar to XGBoost \cite{xgboost}, TencentBoost \cite{tencentboost} and others in that we build distributed quantile sketches of feature values and use those to build histograms, to be used later to find the best split.
In TencentBoost\cite{tencentboost} and PSMART \cite{psmart} full training data is partitioned and loaded in workers' memory, which can be a problem for larger datasets. To address this limitation we instead work on mini-batches, updating quantiles in an online fashion without loading all the data into the memory. To the best of our knowledge, this approach is not implemented anywhere else. 

Each worker loads a small part of data (mini-batch), builds a local quantile sketch, pushes it to PS and then fetches the bucket boundaries that were built at the end of previous iteration. Workers then compute per bucket gradient and Hessian statistics and push them back to the PS. One of the workers, designated as Chief, checks during each iteration if the PS have accumulated enough statistics for the current layer. If so, the Chief starts building the next layer by finding best splits for each of the nodes in the layer. Code that finds the best split for each feature is executed on the PS that have accumulated the gradient statistics for that feature. The Chief receives the best split for every leaf from the PS and grows one layer on the tree. \\
Once the Chief updates the model by adding a new layer, both gradients and quantile statistics become stale. To avoid stale updates, we introduce an abstraction we call StampedResource - a TensorFlow resource with an int64 stamp. Tree ensemble, as well as gradients and quantile accumulators are all stamped resources with such token. When the worker fetches the model, it gets the stamp token which is then used for all the reads and writes to stamped resources until the end of that iteration. This guarantees that all the updates are consistent. This also means that Chief doesn't need to wait for Workers for synchronization, which is important when using preemptible virtual machines. Chief checkpoints resources to disk periodically and workers don't hold any state. So if they are restarted during training, they can simply load a new mini-batch and continue.
\bibliographystyle{splncs03}
\vspace*{-0.65cm}
\bibliography{demo_paper_arxiv}

\end{document}